\documentclass[11pt]{article}
\usepackage{coling2020}
\usepackage{times}
\usepackage{url}
\usepackage{latexsym}

\colingfinalcopy 

\usepackage{times}
\usepackage{latexsym}

\usepackage[T1]{fontenc}

\usepackage[utf8]{inputenc}

\usepackage[scaled=0.8]{beramono}

\usepackage{microtype}
\usepackage{microtype}
\usepackage{url}
\usepackage[T1]{fontenc}
\usepackage{amsmath}
\usepackage{amssymb}
\usepackage{tabularx}
\usepackage{mathtools}
\usepackage{booktabs}
\usepackage{longtable}
\usepackage{tabu}
\usepackage{multirow}
\usepackage{amsfonts}
\usepackage{algorithm}
\usepackage{bbm}
\usepackage{subfigure}
\usepackage[noend]{algpseudocode}
\usepackage[normalem]{ulem}
\usepackage{enumitem}
\usepackage{tikz}
\usetikzlibrary{shapes.geometric}
\usetikzlibrary{patterns}
\usepackage{pgfplots}
\usetikzlibrary{backgrounds}
\usetikzlibrary{matrix,calc}
\def\BState{\State\hskip-\ALG@thistlm}
\usepackage{bm}
\usepackage{xcolor}

\usepackage{tikz-dependency}
\usetikzlibrary{patterns}

\usepackage[T2A,LGR,T1]{fontenc}
\usepackage[utf8]{inputenc}
\usepackage[normalem]{ulem}
\providecommand{\keywords}[1]
{
  \textbf{\textit{Keywords---}} #1
}

\title{Teacher Perception of Automatically Extracted Grammar Concepts for L2 Language Learning }

 \author{Aditi Chaudhary$^\dagger$, Arun Sampath$^\bigtriangleup$, Ashwin Sheshadri$^\bigtriangleup$, Antonios Anastasopoulos$^\ddagger$, Graham Neubig$^\dagger$ \\
  $^\dagger$Carnegie Mellon University, 
  $^\bigtriangleup$Kannada Academy,
  $^\ddagger$George Mason University \\
    \texttt{\{aschaudh,gneubig\}@cs.cmu.edu} \hspace{.5cm} \texttt{antonis@gmu.edu} \\
    \texttt{\{arun.sampath,ashwin.sheshadri\}@kannadaacademy.com}
    }

\date{}

\begin{document}
\maketitle
\begin{abstract}
One of the challenges of language teaching is how to organize the rules regarding syntax, semantics, or phonology of the language in a meaningful manner.
This not only requires pedagogical skills, but also requires a deep understanding of that language.
While comprehensive materials to develop such curricula are available in English and some broadly spoken languages, for many other languages, teachers need to manually create them in response to their students' needs.
This process is challenging because i) it requires that such experts be accessible and have the necessary resources, and ii) even if there are such experts, describing all the intricacies of a language is time-consuming and prone to omission.
In this article, we present an automatic framework that aims to facilitate this process by \emph{automatically} discovering and visualizing descriptions of different aspects of grammar.
Specifically, we extract descriptions from a natural text corpus that answer questions about morphosyntax (learning of \emph{word order, agreement, case marking, or word formation}) and semantics (learning of \emph{vocabulary}) and show illustrative examples.
We apply this method for teaching the Indian languages, Kannada and Marathi, which, unlike English, do not have well-developed pedagogical resources and, therefore, are likely to benefit from this exercise.
To assess the perceived utility of the extracted material, we enlist the help of language educators from schools in North America who teach these languages to perform a manual evaluation.
Overall, teachers find the materials to be interesting as a reference material for their own lesson preparation or even for learner evaluation.
\end{abstract}


\section{Introduction}
\label{sec:intro}

Over the past decade, computer-assisted learning systems (CALL) have gained tremendous popularity, especially during the height of the COVID-19 pandemic when in-person instruction was not possible, leading to the need for user-friendly and accessible learning applications  \cite{li2020covid}.
Even for people learning in traditional classroom settings, the use of CALL systems in tandem  has shown efficacy \cite{callbenefit,barrow2009technology}. 
Typically, the learning materials in these applications must be carefully designed to address the needs of the learning community.
For example, \newcite{network2001guidelines} note that language educators must make effective use of locally relevant expertise and materials when designing curricula, as they embody the cultural heritage of the region.
Because these materials are curated by subject experts, 
this, however, makes curriculum design a challenging process, especially for languages where experts or resources are  inaccessible.

A good curriculum design requires significant time and effort, as this process entails many steps, from designing material for different learning levels, covering different grammar points, finding relevant examples, and even creating evaluation exercises, to name a few.
Additionally, for second language (L2) learning, it is not straightforward to reuse an existing curriculum even in the same language, as the background and requirements of L2 learners could be vastly different from the traditional L1 setting \cite{munby1981communicative}.

With modern computers and techniques, corpus-based methods  \cite{yoon2005investigation} can analyze large text corpora in seconds and find language patterns that can accelerate some of these steps \cite{davies2009exploring}.
Only a handful of languages, in particular English, have a plethora of resources developed for L2 learning, but for most of the world's 7000+ languages, it is a struggle to find even a sufficiently large and good quality text corpus \cite{kreutzer2021quality}, let alone teaching material.
Given this motivation, we explore \emph{to what extent can a combination of natural language processing (NLP) techniques and corpus linguistics assist language education for languages with limited teaching resources?}

Because \emph{corpus linguistics} \cite{biber1998corpus} typically analyzes ``natural text'' i.e. text collected in natural settings with minimal experimental interference, systems based on it have been widely used for learning
 vocabulary \cite{ackerley2017effects,lee2003study}, collocations \cite{chan2005effects,du2022collocation,kanglong2020lexical}, grammar \cite{lin2015data}, L2 writing \cite{yoon2014direct,crosthwaite2020taking}.
However, teachers have mostly used text corpus \cite{bennett2010using,lenko2017training,ma2022teacher} at a very superficial level, i.e. for mainly searching common word patterns, keywords, on popular corpora (e.g. \newcite{davies2008corpus}, \newcite{cobb2002web}, SketchEngine, Skell) to supplement their teaching.
Now, with advances in NLP, we can extract instructional material for more complex linguistic use cases.
For example, we can  automatically answer questions about some \emph{local} aspects of the language, such as \emph{the function of  words} (POS tagging) and \emph{their relations} (dependency parsing).

Inspired by this, we design a framework to aid in language instruction by automatically extracting learning material for ``teachable grammar points'' covering different aspects of grammar, directly from the text corpora of the language of interest.
We define teachable grammar points as individual syntactic or semantic concepts that can be taught to a learner.
For example, with respect to the grammatical aspect of \emph{word order}, a teachable grammar point could be understanding ``how adjectives are positioned with respect to nouns in this language''.%
\footnote{We include lexical semantics of vocabulary selection in ``grammar points'', as vocabulary selection is also an important element of language learning.}
These automatically extracted descriptions also provide human-readable explanations, which detail when one behavior is observed over the other, along with illustrative examples.

We test the utility of this framework through experiments with instructors who teach two Indian languages, Kannada and Marathi, to English speakers.
In addition to our ability to collaborate with in-service  teachers in performing the experiments, both languages also have far fewer pedagogical and NLP resources than English, making them an appropriate testbed for examining the effect of \emph{automatically} extracted  material on the teaching process.
The code and data used to extract the learning materials are publicly available.\footnote{\url{https://github.com/Aditi138/teacher-perception}}
The materials are presented to teachers through an online interface\footnote{\url{https://www.autolex.co}}.



\section{Literature Review}
\label{sec:literature}

\subsection{Corpus-Based Language Pedagogy}
\newcite{jones2015corpus} note that most English textbook writers did not consult a corpus when writing them, but rather relied on their own intuition or followed other textbooks. 
But \newcite{long2000focus} argues that explaining language \emph{only} deductively can be overwhelming to learners and does not expose them as effectively to real-life usage.
Previous research has widely advocated the use of corpora in language teaching \cite{john1991should,flowerdew2011corpora,farr2010can,reppen2010using}, as they expose learners to the ``real'' language.
\newcite{mukherjee2004bridging}  note that few instructors are aware of corpus technology, but after conducting a workshop demonstrating its utility, these instructors realized its potential in teaching.
Therefore, in this work, we collaborate with teachers to examine the possibility of using NLP technology for language instruction.
To our knowledge, the use of such automatically extracted linguistic insights for language teaching has not yet been investigated.
We discuss some of the underlying NLP developments below.


\subsection{Extracting Linguistic Insights Using NLP}
Popular NLP tasks of POS tagging \cite{toutanvoa-manning-2000-enriching} and dependency parsing \cite{kiperwasser-goldberg-2016-simple} provide \emph{local} answers to questions such as ``what is the function (part-of-speech) of a word'' or ``what are the syntactic relations (dependency) between words''.
Over the past decade, these technologies have seen a revolution due to the application of ``deep learning'' \cite{goldberg2017neural}, which can automatically discover patterns in the underlying data.
Notable accuracy gains have been observed for many tasks, such as POS tagging \cite{state}, syntactic \cite{kulmizev-etal-2019-deep}, and morphological analysis \cite{kondratyuk-straka-2019-75}.
Although the amount of in-language syntactically annotated data required to create effective systems is falling rapidly \cite{goldwater-griffiths-2007-fully,agic-etal-2015-bit}, large-scale databases of syntactically annotated data such as the Universal Dependencies (UD) treebanks \cite{nivre-etal-2006-talbanken05,nivre2016universal} have made reasonably sized data available in a wide variety of languages, 
and have set the stage for the use of syntactic analyses in a variety of languages.

While the above-cited works mainly focus on local analysis of individual examples, these analyses can also form the basis for answering more global questions about the grammar as a whole.
For example, a linguist interested in understanding the typical word order (e.g.~subject-verb-object) may perform dependency parsing to identify verbs, subjects, and objects, and aggregate statistics about their relative order.
Prior work \cite{ostling-2015-word,wang-eisner-2017-fine} has looked at such analyses, but does not explain the conditions which trigger one order over the other.
\newcite{chaudhary-etal-2020-automatic} and \newcite{chaudhary2022autolex} have focused on extracting human- and machine-readable grammar patterns to describe morphological agreement, word order, and case marking processes, which were evaluated based on their intrinsic properties.
In this paper, we take a step further by examining the utility of extracted descriptions for language education.

\subsection{Research Questions}
The research question we set out to answer is \emph{whether such automatically derived linguistic insights can be useful for language pedagogy}.
We 
collaborate with in-service teachers in a collaborative design process, where we tailor automatically extracted grammar points to their teaching objectives and practices, and evaluate the utility in their teaching process.
Based on existing materials and prior literature, we identify ``teachable grammar points'' and build on the existing NLP framework, \texttt{AutoLEX} \cite{chaudhary2022autolex}, to extract learning material from the text corpus.
Our study can be further broken down into the following fine-grained questions.
\begin{itemize}
\item How can we extract the learning material for the ``teachable grammar points''?
\item How many of these grammar points extracted are relevant to the learning curriculum?
\item How many of these grammar points are practically usable for language educators to further develop or improve their existing curriculum, and if so, in what ways?
\end{itemize}

\section{Methodology}\label{sec:methodology}
\subsection{Task Overview}
To answer these questions, we identify schools in North America that teach Kannada and Marathi to English speakers. 
Although these languages are spoken primarily in India, a small but significant populace of speakers has emigrated, resulting in a demand to maintain language skills within this diaspora.
Therefore, the primary objective of these teachers is to provide instruction for spoken and written  language to a) preserve and promote the language and culture and b) help speakers communicate with their community. 
Because of these specific objectives, existing textbooks from Indian schools cannot be used as is, as they are based on more \emph{traditional L1 teaching approach} \cite{selvi2018approaches}, where the language is taught from the ground up, from introducing the alphabet, its pronunciation and writing, to each subsequent grammar point.
Teachers have instead adapted the existing material and continue to design new material to suit their requirements.
By using automatically extracted first-pass learning material, teachers could use it as-is or as a supplement to their existing material.
We first conduct a pilot study and collaborative design with two Kannada teachers who are deeply involved in the curriculum design process, which aims to evaluate \emph{quality} and \emph{properties} of the extracted materials.
Next, we conduct a larger study with more volunteer teachers, both in Kannada and Marathi, to evaluate \emph{relevance}, \emph{usability}, and \emph{presentation} of the extracted materials.

\begin{table*}[t]
\small
    \centering
    \resizebox{\textwidth}{!}{
    \begin{tabular}{c|l}
   Grammar Aspect & Teachable Grammar Points \\
    \midrule 
    General Information & What gender values does Marathi show? (e.g. masculine, feminine, neuter) \\
          & Which type of words show these values? (e.g. nouns, verbs) \\
                        & What are some example word usages? \\
    \midrule
    Vocabulary & What words to use for popular categories (e.g. food, animals, etc.) \\
               & What are some adjectives, their synonyms and antonyms? \\
               & Which word to use when? \\ 
    \midrule
    Word Order & Are subjects before or after verbs in Marathi? \\
               & If both, when is subject before and when is it after? \\
    \midrule
    Suffix Usage & What are the common suffixes for Marathi nouns? \\
                 & When should a particular suffix (e.g. -`laa') be used? \\
    \midrule
    Agreement & Do some words need to agree on gender? \\
              & If so, when should they necessarily agree and when they need not?\\

    \end{tabular}
    }
    \caption{Example aspects of the grammar and teachable grammar points covered in our  material.
    }
    \label{tab:lingquestions}
    \vspace{-1em}
\end{table*}

\subsection{Participants and Context}\label{sec:participants}
For Kannada, we work with teachers from the Kannada Academy\footnote{\url{https://www.kannadaacademy.com/}} (KA), which is one of the largest organizations of free Kannada teaching schools in the world, with more than 70 centers.
For the wider study, we recruit volunteer teachers;
KA has 800+ volunteer teachers, of which 12 participated in this study.
For Marathi, there is no central organization as for Kannada, but there are many independent schools in North America. 
We reached out to two such schools, namely Marathi Vidyalay\footnote{\url{https://marathivishwa.org/marathi-shala/}}, in New Jersey, USA, which was established 40 years ago and teaches learners in the age group of 6-15, and Shala in Pittsburgh\footnote{\url{https://www.mmpgh.org/MarathiShala.shtml}}.
Marathi Vidyalay is a small school consisting of seven volunteer teachers, of whom four agreed to participate in the study, while Shala only has one teacher.
All participants are volunteer teachers; teaching is not their primary profession.



\begin{figure*}
    \centering
    \includegraphics[width=\textwidth]{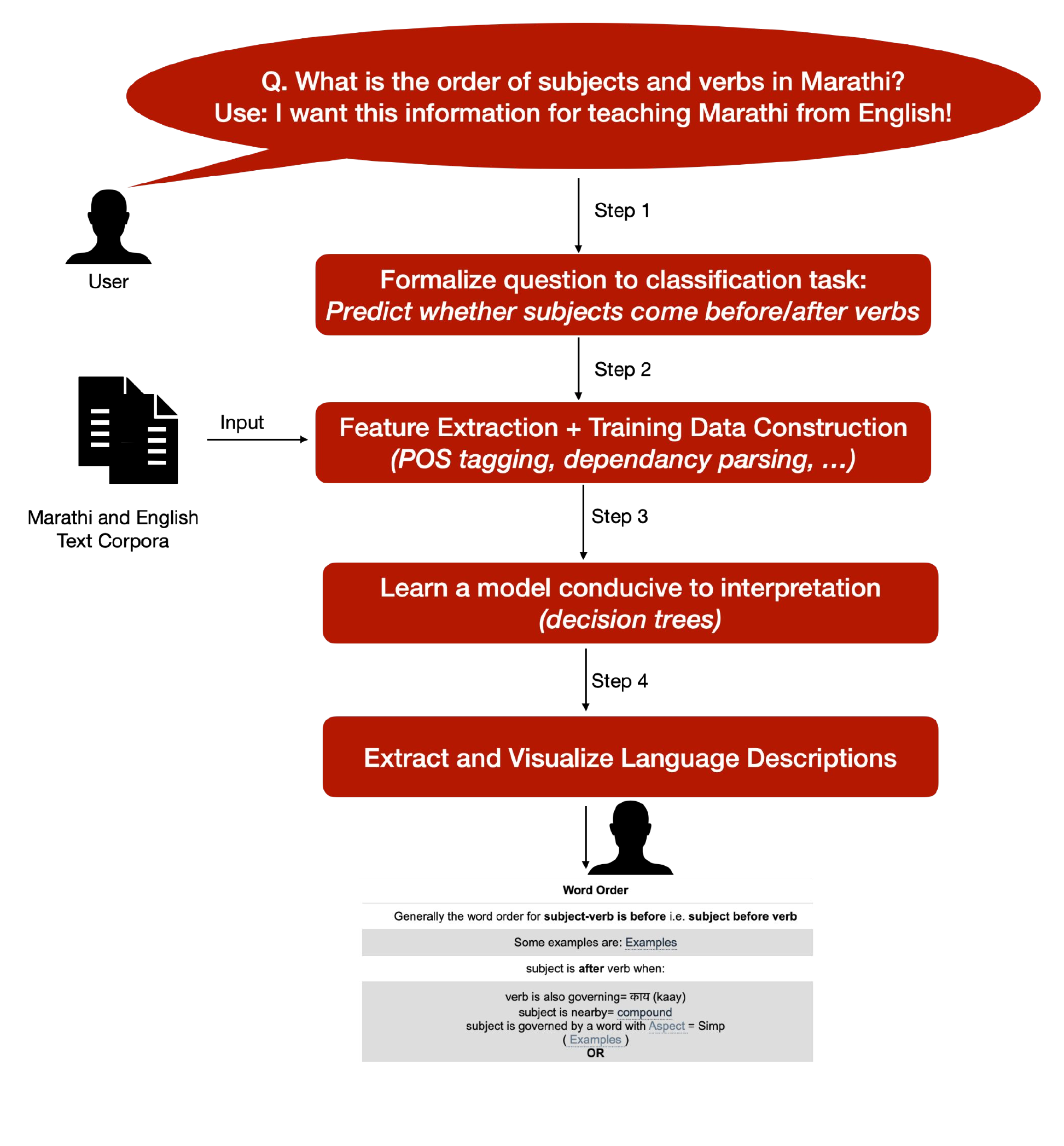}
    \caption{Illustrating the \texttt{AutoLEX} framework.}
    \label{fig:autolex}
\end{figure*}

\subsection{Extracting Learning Materials}\label{sec:extractlearning}
Although language education has been widely studied in literature, there is no one `right' method of teaching.
Several teaching methods have been proposed, and we take inspiration from these to design our materials.
For example, \newcite{doggett1986eight} discuss eight popular  methods, some of which, such as the \emph{Grammar-Translation} method, require learners to translate grammar rules between their L1 and L2 languages, while methods such as \emph{Direct Method, Suggestopedia, Community Language Learning} encourage learning in the L2 language itself.

As noted above, we build on \texttt{AutoLEX} \cite{chaudhary2022autolex} to extract learning material.
\texttt{AutoLEX} takes as input a raw text corpus and produces human- and machine-readable explanations of different linguistic behaviors.
Specifically, given a question (e.g.~``how are objects ordered with respect to verbs in English''), they formalize it into an NLP task and learn an algorithm which extracts not only the common patterns (e.g.~``object is before/after the verb'') but also the conditions which trigger each of them (e.g.~``objects come after verbs except for interrogatives'').
This process is illustrated in Figure \ref{fig:autolex}.
Importantly, for each pattern, illustrative examples and examples of exceptions are extracted from the corpora.

As a first step toward applying \texttt{AutoLEX} to the curriculum design scenario for Kannada and Marathi, we performed an inventory of the aspects taught in existing teaching materials.
We inspected three of the eight Kannada textbooks shared by curriculum designers and identified common grammar points such as identification of syntax categories (e.g.~nouns, verbs, etc.), vocabulary, and suffixes.
In Table \ref{tab:lingquestions}, we show examples of the types of teachable grammar points that we attempt to extract.
Next, we identify a large text corpus in the target languages and apply \texttt{AutoLEX} to extract the answers to these questions. 
We describe the procedure for each grammar point below and refer the reader to \newcite{chaudhary2022autolex} for more details.

\paragraph{Word Order and Agreement}
Both Marathi and Kannada are morphologically rich, with highly inflected words for gender, person, number; morphological agreement between words is also frequently observed.
Both languages predominantly follow subject-object-verb word order, but because syntactic roles are often expressed through morphology, there are often significant deviations from this dominant order.
Therefore, learners must understand both the rules of word order and agreement to produce grammatically correct language. 
In \texttt{AutoLEX} word order and agreement grammar points are extracted by formulating questions, as shown in Table \ref{tab:lingquestions},
and learning a model to answer each question.
To create gold-standard data to train such models, we must first identify the relevant elements (e.g. for word order, subjects, and verbs).
To do so, the corpus must be syntactically annotated with POS tags, morphological analyses, lemmas, and dependency parses.
Next, to discover when the subject is before or after, \texttt{AutoLEX} extracts syntactic and lexical signals from other words in that sentence and uses them to train a classifier.
To obtain interpretable patterns, it is critical to use an interpretable machine learning model, so we opt to use decision trees \cite{quinlan1986induction}, which extract ``if X then Y'' style patterns that can, if presented appropriately, be interpreted by teachers or learners.
Example word order patterns extracted from this model for Marathi are shown in Figure \ref{fig:autolex}.

\begin{figure}
    \centering
    \includegraphics[width=\textwidth]{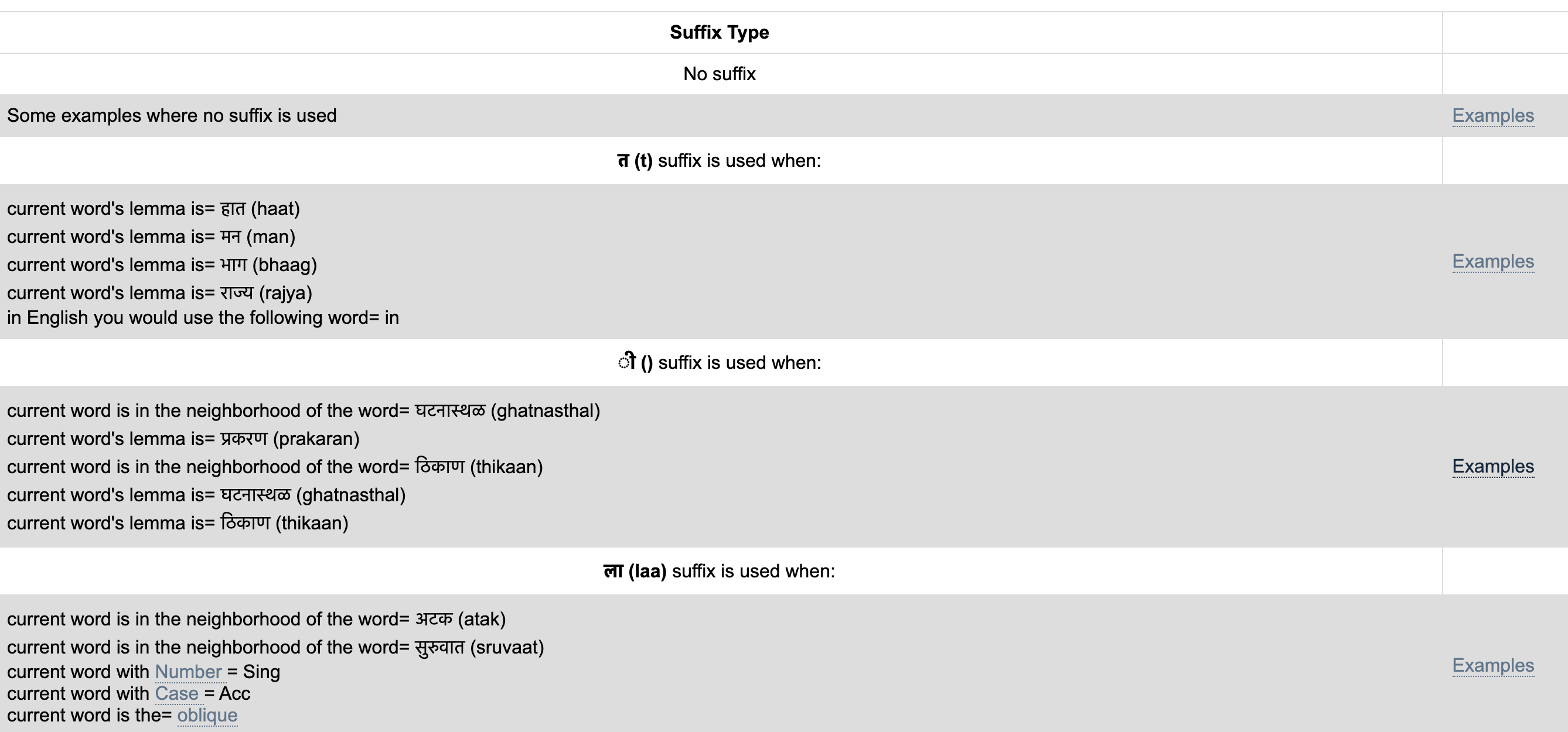}
    \caption{ 
    Marathi suffixes for nouns with their usages.}
    \label{fig:marathiaffix}
\end{figure}
\paragraph{Suffix Usage}
Along with understanding sentence structure, it is equally important to understand how inflection works at the word level, given that these languages are highly inflected.
We first identify the common suffixes for each word type (e.g.~nouns) and then ask ``which suffix to use when''.
Similarly to what we did for word order and agreement, we identify the POS tags and produce a morphological analysis for each word.
To identify the suffix, we then train a model that takes as input a word with its morphological analysis (e.g.~`deshaala,N,Acc,Masc,Sing') and outputs the decomposition (e.g.~`desh + laa').
Next, a classification model is trained for each such suffix (e.g.~`-laa') to extract the conditions under which one suffix is typically used over another (Figure \ref{fig:marathiaffix}).

\begin{figure}
    \centering
    \includegraphics[width=0.7\textwidth]{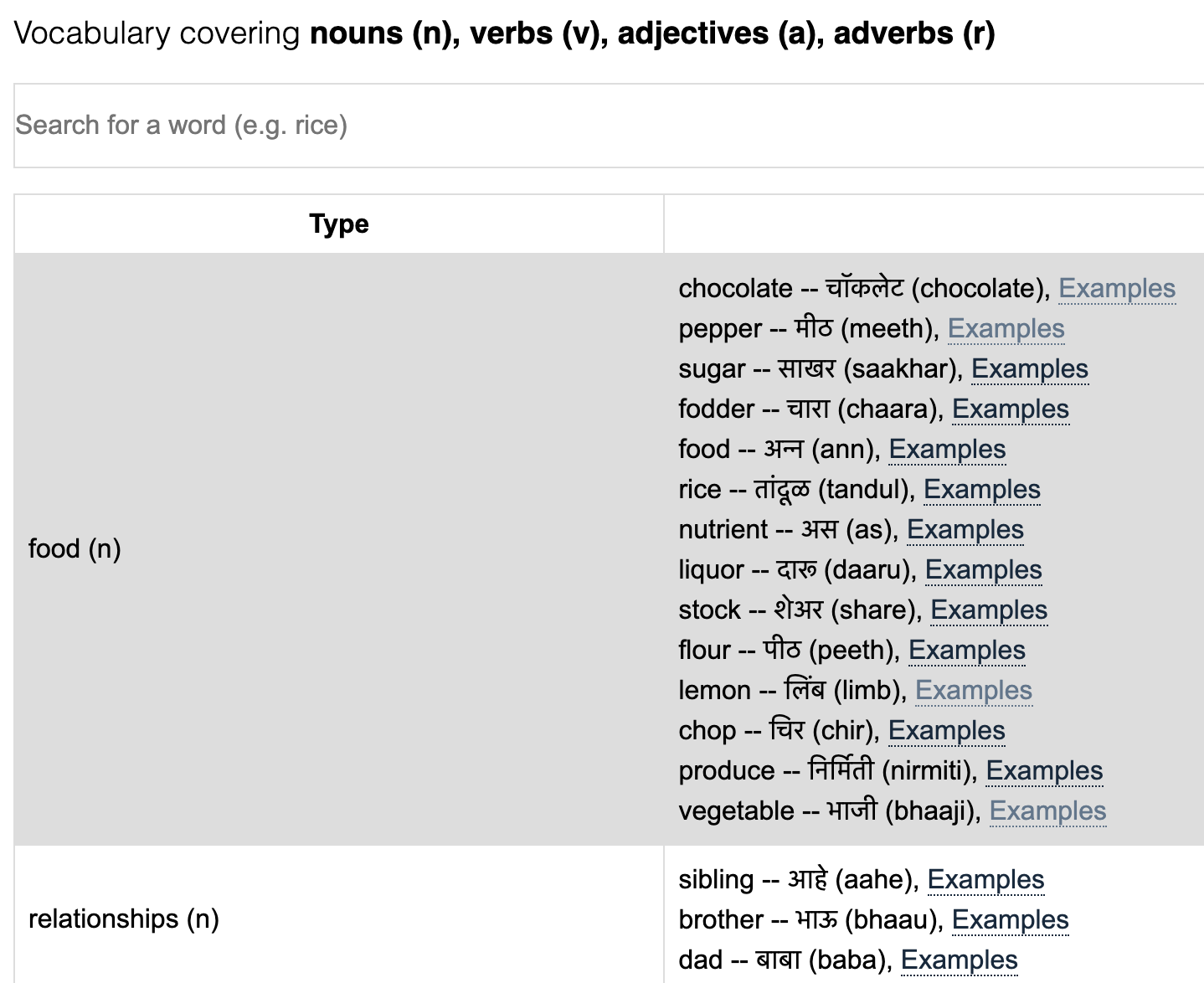}
    \caption{ 
    Marathi words organized by basic categories. Each word contains a link to illustrative examples with their English translations.}
    \label{fig:marathiwordvo}
\end{figure}
\begin{figure}
    \centering
    \includegraphics[width=\textwidth]{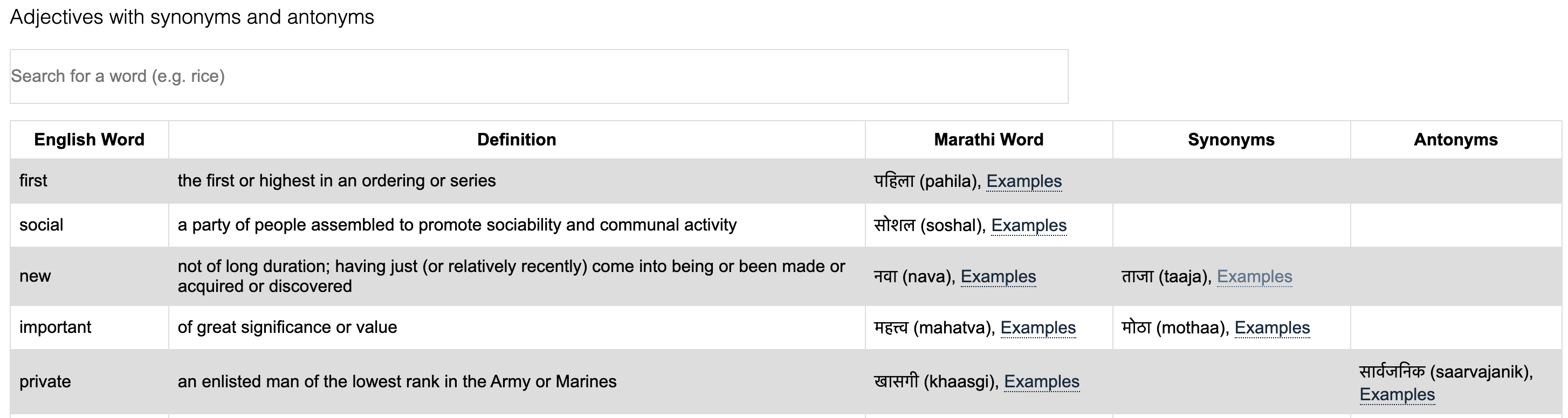}
    \caption{ 
    Marathi adjectives with their definitions, synonyms and antonyms.}
    \label{fig:marathiadjvo}
\end{figure}

\paragraph{Vocabulary}
Vocabulary is probably one of the most important components of language learning \cite{folse2004myths}.
There are several debates on the best strategy for teaching vocabulary; we follow prior literature \cite{groot2000computer,nation2005teaching,richards1999exploring}, which proposes using a mixture of definitions with examples of word usage in context.
Specifically, we organize the vocabulary material around three questions, as shown in Table \ref{tab:lingquestions}.
There are some categories of words where the same L1 (English) word can have multiple L2 translations, with fine-grained semantic divergences (e.g.~`bhaat' and `tandul' both refer to `rice' in Marathi, but the latter refers to raw rice and the former refers to cooked rice).
\newcite{chaudhary-etal-2021-wall} proposed a method for extracting such word pairs, along with explanations on their usage.
Specifically, this starts with parallel data of sentences in English and the L2 language.
Each pair of sentence translations is first run through an automatic word aligner \cite{dou2021word}, which extracts word-by-word translations, producing a list of English words with their corresponding L2 translations.
On top of this initial list, a set of filtration steps is applied to extract those word pairs that show fine-grained divergences.
Training data is then constructed to solve the task of lexical selection, i.e. for a given L1 word (e.g.~`rice') in which contexts to use one L2 word over another (e.g.~`bhaat' vs `tandul').
As mentioned before, \emph{Communicative Approach} \cite{johnson1979communicative} focuses on teaching through functions (e.g.~self-introduction, identification of relationships, etc.) over grammar forms; therefore, we also organize vocabulary around popular categories.
We run a word-sense disambiguation (WSD) model \cite{pasini-etal-xl-wsd-2021} on English sentences, which helps us to identify the word sense for each word in context (e.g.~the word sense `bank.n.02' refers to a financial institution while `bank.n.01' refers to a river edge).
Given the hierarchy of word senses expressed in the lexical resource WordNet \cite{miller1995wordnet}, we can traverse the ancestors of each word sense to find whether it belongs to any of the pre-defined categories (e.g.~food, relationships, animals, fruits, colors, time, verbs, body parts, vehicle, elements, furniture, clothing).
An example of such words extracted for Marathi is shown in Figure \ref{fig:marathiwordvo}.
We also identify popular adjectives, their synonyms, and antonyms, also extracted from WordNet, and present them in a similar format (Figure \ref{fig:marathiadjvo}).
For each word, we also present accompanying examples that illustrate its usage in context, along with its English translations.
For the benefit of users who are not familiar with the script of L2 languages, we automatically transliterate into Roman script using \newcite{Bhat:2014:ISS:2824864.2824872}.

\paragraph{General Information}
In addition to answering these morpho-syntax and semantic questions, we also present salient morphology properties at the language level.
Specifically, from the syntactically parsed corpus of the target language, we hope to answer basic questions such as ``what morphological properties (e.g.~gender, person, number, tense, case) does this language have'', to explain the syntax, as shown in Table \ref{tab:lingquestions}.
These questions were inspired from Kannada textbooks shared by experts, where the textbooks introduce the learner to basic syntax and morphology.
 For each question, we organize the information by frequency, a common practice in language teaching where textbooks often comprise of frequently used examples \cite{dash2008corpus}.

For each grammar point, we use English as the meta-language, as most learners in our study have English as their L1.
In addition to the relevant content, the format in which the material is presented is equally important.
\newcite{smith1981second} outline four fundamental steps involved in language teaching: \emph{presentation}, \emph{explanation}, \emph{repetition} of material until it is learned, and \emph{transfer} of materials in different contexts, which have no fixed order. 
For example, some teachers prefer the presentation of content (e.g.~reading material, examples, etc.) first followed by explanation (e.g.~grammar rules), 
while \newcite{smith1981second} discuss that, for above-average learners, explanation followed by presentation may be preferable.
In our design, we provide both (i.e. rules and examples) without any specific ordering, allowing educators to decide based on their experience and objectives.
By providing illustrative examples from the underlying text at each step, we hope to address the \emph{transfer} step, where learners are exposed to real situations of language use.

\subsection{Evaluating Learning Materials}
\label{sec:survey}
Before presenting the content to volunteer teachers, we first conduct a limited  study to evaluate the appropriateness of the materials.
\paragraph{Quality Study}
Previous work on \texttt{AutoLEX} \cite{chaudhary-etal-2020-automatic,chaudhary2022autolex} conducted a quality study with language experts in multiple languages (English, Greek, Russian, Catalan) that revealed that the rules extracted are of reasonable quality.\footnote{80\% rules extracted for agreement, word order, and case marking were deemed valid for English and Greek, for Russian and Catalan only agreement was evaluated and were deemed 78\% and 66\% valid respectively.}
Therefore, to ensure that we are also achieving a minimal level of quality here, we conduct a similar study only for some Kannada materials with two experts, where we ask them
to evaluate the materials for  word order, word usage and suffix usage.
Specifically,  for word order and suffix usage, we ask  questions regarding 1) whether the rules along with accompanying examples demonstrate the shown concept correctly and 2) if so, whether this material is already covered in their existing  material.
For word usage, we present the experts with the extracted words with their English translations and ask them how many of these  are correct.%
\footnote{\newcite{chaudhary-etal-2021-wall} have already evaluated the efficacy of the extracted rules for word usage in a true learning setup for Greek and Spanish, and therefore we focus on evaluating only the word pairs and not the rules.}

\begin{table*}[t]
    \centering
    \resizebox{\textwidth}{!}{
    \begin{tabular}{c|l|l}
    \textbf{Type} & \textbf{Question} & \textbf{Answer Choices}  \\
    \midrule 
    Relevance & \textbf{1.} What percentage of the materials presented & 0-100\% \\
              & in the tool cover  & \\
              & existing curriculum requirements? & \\
   
    \midrule
    Utility & \textbf{2.} How likely are you personally inclined to use & 3: Highly likely \\
            & this tool in your lesson planning or teaching? & 2: Likely \\
            & &  1: Not likely \\
            & & \\
           & \textbf{2.1.} If likely, for what purpose do you foresee  & a. For lesson preparation, knowledge \\
           & this being used? & b. For evaluating students \\
           & (multiple answers can be selected): & c. Present this to the students for self-exploration \\
            & & d. Other (please specify the reason) \\
            & & \\
           & \textbf{2.2.} If likely,  what aspects would you use: & a. The general concept  introduced by the material \\
           &  &  b. The rules which are described in the table \\
           & (multiple answers can be selected): &  c. Illustrative examples that accompany the rule \\
           & & d. Other (please specify the reason) \\
           & & \\
           & \textbf{2.3.} if NOT likely,  why? & a. material outside the scope of current curriculum \\
           &  &  b. material was unclear and needs improvement  \\
           & (multiple answers can be selected): &  c. material is already covered by existing curriculum \\
           & & d. Other (please specify the reason) \\
    \midrule
    Presentation &  \textbf{3.} How did you find the tool? & 3. Very easy to use and navigate \\
                 &  & 2. Somewhat easy to use, but took some time to get used to \\
                 &  & 1. Difficult to use \\
    \midrule
    Feedback & \multicolumn{2}{l}{\textbf{4.1} What did you like about the tool or the learning materials?}  \\
             & \multicolumn{2}{l}{\textbf{4.2} What did you not like about the tool?}  \\
             & \multicolumn{2}{l}{\textbf{4.3} What would you like to improve in the tool?}  \\
    \end{tabular}
    }
    \caption{Usability study: Questions posed to the in-service teachers for evaluating the learning materials on relevance, utility and presentation.
    This set of questions is asked for each grammar concept.}
    \label{tab:questionnaire}
    \vspace{-1em}
\end{table*}
\paragraph{Perception Study}
To answer the research question of whether materials are practically usable and, if so, with regard to what aspects, we analyze a broader set of teachers' \emph{perception} of the materials presented.
We conduct this study with instructors of both Kannada and Marathi. 
Through this study, we hope to understand \emph{relevance},  \emph{utility} and \emph{presentation} of the materials.
This study is conducted in three parts; first, a 30--60 minute  meeting is conducted for the teachers, in which we introduce the tool, the different grammar points covered in it, and how to navigate the online interface.
Teachers have one week to explore the materials.
Finally, all teachers receive a questionnaire that requires them to assess the relevance, utility and presentation, as shown in Table \ref{tab:questionnaire}.
We ask this set of questions for each grammar aspect (i.e. general information, vocabulary, suffix usage, word order, and agreement), along with their overall feedback which is more open-ended.

\section{Experimental Setting}
In this section, we describe the details of the data and models used to extract the learning materials.

\paragraph{Data}
Since our goal is to create teaching material for learners having English as L1, 
we use the parallel corpus of Kannada-English and Marathi-English from the dataset \textsc{Samanantar} \cite{ramesh2022samanantar}.
This consists of 4 million Kannada sentences and 4 million Marathi sentences with their respective English translations, and covers text from a variety of domains such as news, Wikipedia, talks, religious text, movies.
Of these genres, the corpus has a particularly large amount of newspapers and legal proceedings, and thus consists of more formal and traditional language than typically appears in textbooks.

\paragraph{Model}
As mentioned in Section \ref{sec:extractlearning}, the first step in the extraction of materials is to parse sentences for POS tags, morphological analysis and dependency parsing.
To obtain this analysis for our corpus, we use an automatic parser \textsc{Udify} \cite{kondratyuk-straka-2019-75} that jointly predicts POS tags, lemma, morphology and dependency tree over raw sentences.
To train a parser for Marathi and Kannada, we use the training data collected by IIIT-Hyderabad\footnote{\url{https://ltrc.iiit.ac.in/showfile.php?filename=downloads/kolhi/}}, which is annotated in the Paninian Grammar Framework \cite{bhat2017hindi}.
However, the \textsc{Udify} model requires training data in the UD annotation scheme \cite{mcdonald-etal-2013-universal}, so we follow \newcite{tandon-etal-2016-conversion} to convert between the two formats to obtain POS tags, lemmatization and morphological analysis. 
However, this converted data does not have dependency information.
To obtain dependency data, we first train the \textsc{Udify} model in a related language (Hindi) and apply it directly to the converted data above.
We then train a new model on this converted data and augment it with the Hindi data as well, and apply the resulting model on the 4 million Marathi and Kannada raw sentences.
We follow the same modeling setup as \newcite{chaudhary-etal-2021-wall} and \newcite{chaudhary2022autolex} to extract the patterns, explanations and accompanying examples.
For suffix usage, we additionally train a morphology decomposition model \cite{ruzsics-etal-2021-interpretability} which breaks a word into its lemma and suffixes.

\section{Results}
In this section, we present the results of the quality and perception studies.
In addition to human evaluation, we follow the strategy of \newcite{chaudhary2022autolex} to automatically evaluate the quality of extracted descriptions.
This is done by applying the learnt model on a held-out set of sentences and observing the prediction accuracy for each linguistic question.
We apply the same evaluation protocol for word order, suffix usage and agreement, and report the results in Table \ref{tab:autolex_evaluation}.
We can see that in most cases, the rules extracted by the model outperform the respective baselines, suggesting that the model is able to extract decent first-pass rules, with 98\%  accuracy for Kannada word order, 48\% for agreement, 85\% for suffix usage, 68\% for vocabulary, 98\% for Marathi word order, 61\% for agreement, 85\% for suffix usage and 70\% for vocabulary.
\begin{table*}[t]
    \begin{center}
    \resizebox{\textwidth}{!}{
    \begin{tabular}{c|c||c|c||c|c}
    \toprule
    & & \multicolumn{2}{c||}{Kannada} & \multicolumn{2}{c}{Marathi} \\
    \textbf{Grammar Concept} & \textbf{Type} & \textbf{\textsc{AutoLEX}} & \textbf{baseline} & \textbf{\textsc{AutoLEX}} & \textbf{baseline}  \\
    
    \midrule
    Word Order & subject-verb & \textbf{97.02} & 96.97 & \textbf{97.8} &	97.7\\
               & object-verb & \textbf{99.11} & 99.06 & \textbf{97.89} &	96.78\\
                & numeral-noun &\textbf{98.63} & 98.36 & 99.54 &	99.54\\
               & adjective-noun & 99.92 & 99.92 & - & - \\
               & noun-adposition & 99.14 & 99.14 & - & - \\
              
    \midrule
    Agreement & Gender & \textbf{71.87} & 65.69 & 61.11 & \textbf{81.44}\\
              & Person & 24.73 &		\textbf{25.16} & - & - \\
    \midrule
    Suffix Usage &  NST	& \textbf{91.58} &	50 & \textbf{90.44} & 93.77\\
                 & NUM	& \textbf{85.2}	 & 	82.63 & 85.91 &	\textbf{93.61}\\
                 & NOUN	& \textbf{78.61} &	39 & \textbf{70.23} & 67.8 \\
                 & PRON	& \textbf{87.13} &	58.03 & \textbf{75.66} & 65.07 \\
                 & PART	& \textbf{94.73} &	89.35 & \textbf{90.58} & 76.77\\
                 & ADJ	& \textbf{87.74} &	66.82 & \textbf{87.55} & 83.83 \\
                 & VERB	& \textbf{63.19} &	30.52 & \textbf{78.44} & 65.87 \\
                 & PROPN & \textbf{74.57} &	46.68 & 65.6 &	\textbf{71.19} \\
                 & SCONJ &	\textbf{96.85} & 64.6 & \textbf{97.59} & 86.9 \\
                 & DET	& \textbf{99.53} &	61.83 & \textbf{83.91} & 81.71 \\
                 & AUX	& \textbf{76.92} &	38.46 & \textbf{92.8} &	81.57 \\
                 & ADV	& \textbf{75.19} &	37.27 & \textbf{86.84} & 65.89 \\
                 & ADP	& \textbf{93.55} &	76.43 & \textbf{97.12} & 67.63 \\
    
    \midrule
    Vocabulary & Semantic Subdivisions & \textbf{68.68} & 58.48 & \textbf{70.58} & 56.26 \\
    \end{tabular}
    }
    \caption{Automated evaluation results for learning materials extracted for each grammar concept}
   \label{tab:autolex_evaluation}
     \end{center}
\end{table*}


			




\subsection{Quality Study Results}
First, we present the result of our study with Kannada experts, where we asked them to evaluate the materials for word order, suffix usage and word usage.

\paragraph{Vocabulary}
We present both experts with an automatically generated list of pairs of 100 English-Kannada words, where one English word has multiple Kannada translations for which the extraction procedure has identified fine-grained divergences.
Both experts found 80\% of the word pairs to be valid, according to criterion of the words showing different usages.
For example, for `doctor', the model discovered four unique translations, namely `vaidya, vaidyaro, daktor, vaidyaru' in Kannada which the expert found interesting for teaching as they demonstrated fine-grained divergences, both semantically and syntactically. For instance, `vadiya' is the direct translation of `doctor', whereas `daktor' is the English word used as-is, `vaidyaro' is the plural form of doctors and `vaidayaru' is a formal way of saying a doctor.
Regarding usability, the expert mentioned that currently in their curriculum there is no good way of handling synonyms or such fine-grained divergences; they said:
\begin{quote}
    ``\emph{given that these word pairs have been extracted from natural text, its interesting to see that there are certain word senses which are so frequently used in the real world which currently we haven't covered in our lesson but are we are now thinking of adding.}''
\end{quote}
Another point they mentioned was that teachers struggle to come up with different examples to illustrate word usages, so the accompanying examples we present are ``extremely useful''.

\paragraph{Word Order}
For word order, the rules for the subject-verb and object-verb order were evaluated by experts.
For subject-verb, six grammar rules explaining the different word order patterns were extracted (4 explaining when the subject can occur both before and after the verb, 2 rules informing when subjects occur after, and 1 showing the default order of ``before'').
Of the six rules, experts found three to be valid patterns.
For object-verb word order, of the six rules extracted by the model, experts marked that 2 rules precisely captured the patterns, while one rule was too fine-grained.
Interestingly, all these rules which were deemed valid were the ones which showed non-dominant patterns.
The rules marked as invalid were invalid because the syntactic parser that generated the underlying syntactic analyses incorrectly identified the subjects/objects.
Such errors are expected given that there is not sufficient quantity/quality of expertly annotated Kannada syntactic analyses available to train a high-quality parser.
However, we would argue that these results are still encouraging because i) despite imperfect syntactic analysis, the proposed method was nonetheless able to extract several interesting counter-examples to the dominant word order, and ii) as mentioned in Section \ref{sec:literature}, further improvements in the underlying parsers for low-resource languages may be expected through active research.

The experts also noted that our material for word order is not covered in any textbook they use, but mentioned that it is suitable for textbooks for advanced learners.
Along with the rules, the material also presents illustrative examples that demonstrate these rules in real-world contexts, and the experts found this to be the most beneficial.
In the words of the expert --
\begin{quote}
     ``\emph{the examples could become exercise material to evaluate learners. They could also be used as inspiration to create simpler examples, as some examples had some words  omitted for brevity''}.
\end{quote}
One challenge that experts pointed out is that most volunteer teachers 
are not trained in formal linguistics, meaning that some might not use such terminology of ``subject, object'' in classroom teaching.
However, experts did mention that the material could also be beneficial for teachers themselves to think more about the structure of Kannada, as some of these non-dominant word order usages are interesting and often learners do ask questions  about such exceptions.

\paragraph{Suffix Usage}
We extract the different suffixes used for each word type (e.g.~nouns, adjectives, etc.) but in the interest of time asked experts to evaluate only the suffixes extracted for nouns and verbs.
Of the 18 noun suffixes, 7 were marked as valid, 2 suffixes were not suffixes in traditional terms but arise due to ``sandhi'' i.e. transformation in the characters at morpheme boundaries.
Similarly, for verb suffixes, 53\% (7/13) were marked as valid.
Experts mentioned that understanding suffix usage is particularly important in Kannada, as it is an agglutinative language with different affixes for different grammar categories.
They identified that although some suffixes (e.g.~--ga.Lu, --i.s) were covered in their existing textbooks, the examples shown in our extracted materials could still be helpful in teaching.

\subsection{Perception Study Results}
In this section, we discuss the results of the perception study that aims to understand how Marathi and Kannada teachers perceive the material presented for their teaching process.

\begin{table*}[t]
    \centering
    \resizebox{\textwidth}{!}{
    \begin{tabular}{c|c|l|l|l}
    \textbf{Grammar Concept} & \textbf{Relevance} & \multicolumn{2}{c|}{\textbf{Utility}} & \textbf{Presentation}  \\
    & \textbf{\% of relevant} & \textbf{\% of teachers} & \textbf{\% of teachers } & \textbf{\% of teachers} \\
    & \textbf{curriculum covered} & \textbf{likely to use} & \textbf{that would use for} & \textbf{found this \rule{1cm}{0.15mm} to navigate} \\
    \toprule 
    General Information & 62.1\% & highly likely: 8.3\% & \textbf{lesson prep: 81.8\%} & very easy: 33.3\% \\
                        &     & \textbf{likely: 83.3\%} & student exploration: 54.5\% &  \textbf{somewhat easy: 58.3\%} \\
                        & & not likely: 8.3\% & student evaluation: 10\% &  difficult: 8.3\% \\
    \midrule
    Vocabulary & 67.5\% & highly likely: 33.3\% & \textbf{lesson prep: 72.7\%} & 
    very easy: 36.3\% \\
               &      & \textbf{likely: 58.3\%} & student exploration: 72.7\% & \textbf{somewhat easy: 58.3\%} \\
               & & not likely: 8.3\% & student evaluation: 45.5\% & difficult: 8.3\% \\
    \midrule
    Suffix Usage & 52.5\% & highly likely: 9.1\% & \textbf{lesson prep: 77.8\%} & very easy: 36.4\% \\
     &     & \textbf{likely: 72.7\%} & student exploration: 55.6\% &  \textbf{somewhat easy: 63.6\%} \\
     & & not likely: 18.2\% & student evaluation: 33.3\% & difficult: 0\% \\
    \midrule
    Word Order & 66\% & highly likely: 10\% & \textbf{lesson prep: 88.9\%} & very easy: 27.3\% \\
     &     & \textbf{likely: 70\%} & student exploration: 44.2\% &  \textbf{somewhat easy: 72.7\%} \\
     & & not likely: 20\% & student evaluation: 22.2\% & difficult: 0\% \\
    \midrule
    Agreement & 53.75\% & highly likely: 20\% & \textbf{lesson prep: 77.8\%} & very easy: 36.4\% \\
     &     & \textbf{likely: 60\%} & student exploration: 44.4\% &  \textbf{somewhat easy: 45.5\%} \\ 
     & & not likely: 20\% & student evaluation: 22.4\% & difficult: 18.2\% \\
    \end{tabular}
    }
    \caption{Perception study results for Kannada. 12 teachers participated in this study}
    \label{tab:questionnaireresults}
    \vspace{-1em}
\end{table*}
\subsubsection{Kannada Results and Discussion}
12 teachers with varying levels of teaching experience participated in this study.
Three teachers had less than three years of experience, four teachers had between 3-10 years, and the remaining four  had 10+ years of experience.
Three teachers teach only beginners, while others have experience teaching higher levels as well.
All teachers have used some online tools, but mostly for creating assignments  for the learners (e.g.~through Google Classroom, Kahoot\footnote{\url{https://kahoot.com/}}, Quizlet\footnote{\url{https://quizlet.com/}}), or conducting classes (e.g.~Zoom).
Some teachers have also referred to YouTube videos and online dictionaries such as Shabdkosh\footnote{\url{https://www.shabdkosh.com/dictionary/english-kannada/}} as reference materials.
However, none had used online tools similar to \texttt{AutoLEX}, which in addition to vocabulary explains the concepts of syntax and morphology with illustrative examples.
As described in Section \ref{sec:survey}, we ask questions centered on the utility, relevance, and presentation  for each grammar concept covered by the tool.
We report individual results in Table \ref{tab:questionnaireresults}.
Furthermore, one teacher, whom we will refer as T1, referred to the \texttt{AutoLEX} materials in one of their lessons which we also discuss below.

\paragraph{Relevance}
In general, we see that teachers, on average, find 45--60\% of material presented as relevant to their existing curriculum.
This is notable given that the underlying  corpus is not specifically curated for language teaching and contains rather formal language.
The teachers note that especially for beginners they prefer starting with simpler and more conversational language style, but for advanced learners this would be very helpful, in their own words--
\begin{quote}
    \emph{``The examples are well written, however, for the beginners and intermediates, this might be too detailed. 
    The corpus could be from a wider data source. The use of legal and court-related terms is less commonly used in day-to-day life.
Advanced learners will certainly benefit from this.''}
\end{quote}

\paragraph{Utility}
We find that for all grammar concepts, most teachers expressed that they were likely to use the materials for lesson preparation.
In fact, the teacher, T1, in fact used our materials to teach suffix usage to an adult learner and said that--
\begin{quote}
    \emph{``I used this tool to teach an American adult who takes private lessons from me and found it helpful in addressing her grammar questions. 
    I liked how it was clearly segregated i.e. the suffixes for nouns vs proper-nouns and how it is different from one another. It is definitely great tool to refer for adults but again the vocabulary is perfect to improve  written skills than the spoken language''.}
\end{quote}
Some teachers also mentioned that they could present the material to students for self-exploration, and about 70\% teachers noted that it would be especially helpful for vocabulary learning.
When asked what aspects of the presented material they would consider using, all teachers said that they would, in particular, use the illustrative examples for all sections except for the word order and agreement sections.
Within the examples, T1 pointed out that it would be more beneficial if the examples could start with simple sentences with the ability to slowly add complex elements to it, they said--
\begin{quote}
   \emph{``The examples used should start with simple sentences, and as per the appetite, teachers should be able to increase the complexity.''}
\end{quote}
For agreement and word order  sections, although they liked the general concepts presented in the material (for example, the non-dominant patterns shown under each section), 88\% of the teachers felt that the material covered advanced topics outside the current scope of their teaching.
Although the quality evaluation of the rules was not part of this study, teachers noted that if the accuracy of the rules, particularly for suffix usage, could be further improved, they could foresee this tool being used in classroom teaching, as suffixes are essential in Kannada.

\paragraph{Presentation}
In terms of presentation of the materials, we can see from the Table \ref{tab:questionnaireresults}, all teachers found them easy to navigate through, although it took some getting used to.
This is expected given that the teachers spent only a few hours (5-6) over the course of a week exploring all materials.
Additionally, the meta-language used to describe the materials consisted of formal linguistic jargon (for example, most teachers were unfamiliar with the term `lemma') 
and some teachers noted that:
\begin{quote}
    \emph{``Tool is great and provides clues and ideas for teaching.
    This is a very vast material; it is a maze where you can easily get lost.
    The idea of pattern recognition is a very natural way of learning for children who relate to audio and visual patterns to grasp concepts. So, this tool helps a lot.''}
\end{quote}



\subsubsection{Marathi Results and Discussion}
Compared to the Kannada study, only five teachers participated in the Marathi study.
These teachers volunteer in small schools that teach mainly at the beginner level with a few intermediate learners.
We report the individual results in Table \ref{tab:questionnaireresultsmarathi}.

\paragraph{Relevance} 
Teachers find only 10--15\% of the materials presented as relevant to their existing curriculum.
This is much less than what the Kannada teachers reported, probably because the Marathi schools' primary focus is teaching beginners.
For beginners, teachers begin by introducing the alphabet, simple vocabulary, and sentences. 
In our tool, currently we do not curate the material according to learner age/experience, and we have extracted the learning materials from a publicly available  corpus which comprises of news articles that are not beginner-oriented.
In fact, the teachers do mention that--
\begin{quote}
    ``\emph{We focus on varnamala i.e. letters, need to figure out how to use material for 6-7 years old students as
the basics are there, but many of the words  are from core Marathi newspaper based language, which is very difficult for kids to grasp. They need simpler words and sentences to effectively understand the words and build vocabulary. There should be some age and language skill-based approach to this learning.}''
\end{quote}
\begin{table*}[t]
    \centering
    \resizebox{\textwidth}{!}{
    \begin{tabular}{c|c|l|l|l}
    \textbf{Grammar Concept} & \textbf{Relevance} & \multicolumn{2}{c|}{\textbf{Utility}} & \textbf{Presentation}  \\
    & \textbf{\% of relevant} & \textbf{\% of teachers} & \textbf{\% of teachers } & \textbf{\% of teachers} \\
    & \textbf{curriculum covered} & \textbf{likely to use} & \textbf{that would use for} & \textbf{found this \rule{1cm}{0.15mm} to navigate} \\
    \toprule 
    General Information & 15\% & highly likely: - & \textbf{lesson prep: 100\%} & very easy: - \\
                        &     & likely: 40\% & student exploration: 50\% &  \textbf{somewhat easy: 80\%} \\
                        & & \textbf{not likely: 60\%} & student evaluation: - &  difficult: 20\% \\
    \midrule
    Vocabulary & 16\% & highly likely: - & \textbf{lesson prep: 100\%} & 
    very easy: - \\
               &      & likely: 40\% & student exploration: 50\% & \textbf{somewhat easy: 100\%} \\
               & & \textbf{not likely: 60\%} & student evaluation: 50\% & difficult: - \\
    \midrule
    Suffix Usage & 9\% & highly likely: - & \textbf{lesson prep: 100\%} & very easy: - \\
     &     & likely: 40\% & student exploration: 50\% &  \textbf{somewhat easy: 100\%} \\
     & & \textbf{not likely: 60\%} & student evaluation: - & difficult: - \\
    \midrule
    Word Order & 8\% & highly likely: - & \textbf{lesson prep: 100\%} & very easy: - \\
     &     & likely: 40\% & student exploration: 50\% &  \textbf{somewhat easy: 80\%} \\
     & & \textbf{not likely: 60\%} & student evaluation: - & difficult: 20\% \\
    \midrule
    Agreement & 5\% & highly likely: - & \textbf{lesson prep: 100\%} & very easy: - \\
     &     & likely: 40\% & student exploration: 50\% &  \textbf{somewhat easy: 80\%} \\ 
     & & \textbf{not likely: 60\%} & student evaluation: - & difficult: 20\% \\
    \end{tabular}
    }
    \caption{Perception study results for Marathi. 5 teachers participated in this study}
    \label{tab:questionnaireresultsmarathi}
    \vspace{-1em}
\end{table*}

\paragraph{Utility}
Similarly to the Kannada findings, all teachers noted that they would likely use the materials for lesson preparation. 
Some teachers also said that they could provide the materials to advanced students for self-exploration, to encourage them to explore the materials and ask questions.
Similarly to the Kannada study, the teachers found the illustrative examples to be of the most utility, as they demonstrate a variety of usage.
However, they did note that because the underlying corpus was too restricted in genre, they would benefit more from applying this tool to their curated set of stories, which are written in age-appropriate language.


\paragraph{Presentation}
All teachers found the materials somewhat easy to navigate and, similarly to Kannada teachers, they mentioned that it did require some time to understand the format.
Some teachers said that currently the material is too content heavy and not visually engaging, if the presentation could be improved along those aspects, it would make the tool more inviting.
Another common piece of feedback about the presentation was regarding content organization -- teachers felt it would be more helpful if the content could be broken into sub-sections with not all technical details added in a single layer.
For instance, they said--
\begin{quote}
    ``\emph{There is a lot of technical details with various parts of speech, subject, object in one go. Perhaps it could be broken down into sub-sections for someone new to the language,  it would be helpful to have clarity on the rules in simpler way followed by simpler examples.
}''
\end{quote}

\section{Conclusion and Next Steps}
In general, from the above discussion, we see that both the Marathi and Kannada findings have some common themes, such as --
\begin{itemize}
    \item Teachers find the selected grammar points relevant to their teaching, which highlights the importance of a collaborative design by involving the stakeholders in the process; however, all note that the content is more suitable for advanced learners.
    
    \item Among the different features, teachers find the illustrative examples to be most useful, especially for understanding the non-dominant linguistic behaviors or the exceptions to general rules.
    
    \item Teachers find this overall effort promising, as this tool can be applied to a corpus of their choice, which is more suited for the learning experience or requirements. 
\end{itemize}

Based on the study, we provide recommendations for the design of future systems. 
\begin{itemize}
    \item Currently, a major limitation of the tool, as noted by the teachers, is that the content is not organized by learner age/experience.
    A next step would be to invite teachers to organize the content by each level, taking the learner incrementally through the complexities of language.
    
    \item For beginner learners, language properties are built through engaging stories with little use of formal grammar terms.
    Therefore, using simpler meta-language to explain the grammar points and including engaging content would be a worthwhile addition.
\end{itemize}



\section*{Acknowledgements}
We are grateful to Charles Perfetti and Lin Chen from the University of Pittsburgh, for the illuminating discussions on L2 acquisition, during early stages of this work.
We thank all teachers, without whom this work would not have been possible or meaningful.
Specifically, we are grateful to the teachers of the Kannada Academy-- Arun Sampath, Ashwin Sheshadri, Manasa Kashi, Mukta Hendi, Sunita Sundaresh, Aravind Gangaiah, Shashi Basavaraju, Madhu Rangappagowda, Gayathri Hebbar, Shruthi A, Naina Sharma, Gowri Gudi and P Tantry.
We are also grateful to the Marathi Vidyalay, New Jersey teachers-- Sudhir Ambekar, Aparna Potdar, Sujata Kulkarni, Varsha Joshi and Archana Kakirde and the Marathi Shala teacher from Pittsburgh-- Pranati Talnikar.
We also thank Sunanda Tumne, Komal Chaukkar, and
Sona Bhide of the Bruhan Maharashtra Mandal (BMM) Shala for providing initial feedback on the design of the interface.
We also thank Pruthwik Mishra and Dipti Misra from IIIT-Hyderabad for sharing the Marathi and Kannada treebanks for training the parser.
This work is sponsored by the Waibel Presidential Fellowship and the
National Science Foundation under grants 1761548 and 2125466.

\bibliographystyle{coling}
\bibliography{acl}


\end{document}